\pdfoutput=1

\documentclass[11pt]{article}

\usepackage{ACL2023}

\newcommand\blfootnote[1]{%
  \begingroup
  \renewcommand\thefootnote{}\footnote{#1}%
  \addtocounter{footnote}{-1}%
  \endgroup
}

\usepackage{times}
\usepackage{latexsym}
\usepackage{graphicx}
\usepackage{url}
\usepackage{linguex}
\usepackage{blindtext}

\usepackage[T1]{fontenc}

\usepackage[utf8]{inputenc}

\usepackage{microtype}

\usepackage{inconsolata}

\usepackage[english]{babel}

\title{Different Tokenization Schemes Lead to Comparable Performance in Spanish Number Agreement}

\author{Catherine Arnett*$^{1}$, Pamela D. Rivière*$^{2}$, Tyler A. Chang$^{2,3}$, Sean Trott$^{2}$ \\
        $^1$Department of Linguistics, \\
        $^2$Department of Cognitive Science, \\
        $^3$Halıcıoğlu Data Science Institute \\
        UC San Diego \\
         \texttt{\{ccarnett, pdrivier, tachang, sttrott\}@ucsd.edu}\\
  }

\begin{document}
\maketitle

\begin{abstract}
The relationship between language model tokenization and performance is an open area of research. Here, we investigate how different tokenization schemes impact number agreement in Spanish plurals.
We find that morphologically-aligned tokenization performs similarly to other tokenization schemes, even when induced artificially for words that would not be tokenized that way during training.
We then present exploratory analyses demonstrating that language model embeddings for different plural tokenizations have similar distributions along the embedding space axis that maximally distinguishes singular and plural nouns.
Our results suggest that morphologically-aligned tokenization is a viable tokenization approach, and existing models already generalize some morphological patterns to new items. However, our results indicate that morphological tokenization is not strictly required for performance.

\end{abstract}

\section{Introduction}

In natural language processing (NLP) pipelines, \textbf{tokenizers} segment unstructured text into smaller, discrete constituents (``tokens'') for further processing.\blfootnote{*Equal contribution.}
Importantly, different tokenizers can incur performance and efficiency trade-offs.
Assigning a unique token to each word in a corpus may lead to high-precision semantic representations, but the resulting models might be less robust to unseen words and require more computational resources.

Most existing tokenizers allow words to be decomposed into subword tokens \citep{sennrich-etal-2016-neural,kudo-richardson-2018-sentencepiece}. They can do so along morphological boundaries (e.g. \textit{books} to [`book', `\#\#s']), but this behavior is not guaranteed.
Segmenting words into their lemmas and morphemes might simultaneously allow models to more robustly learn morphosyntactic patterns, more efficiently represent such patterns, and better generalize to novel words. (An analogous question concerning the storage of whole words vs. learning generalizable rules exists within human psycholinguistics research, e.g., \citealp{ullman2016declarative}). 

Thus, the present work evaluates the effect of three types of plural noun tokenization in Spanish---single-token plurals, morphemically-tokenized plurals, and non-morphemically-tokenized plurals---in the context of a masked article prediction task (\S\ref{sec:study}). 

We find that tokenization schemes are differentially successful, although the effect is small, and article agreement accuracy is high across all tokenization types. Artificial tokenization schemes, where we coerce an initially single-token or non-morphemically-tokenized plural into a morphemic representation, leads to successful task performance, but does not improve performance beyond the original tokenization scheme. In an exploratory analysis, we compare singular and plural form embeddings across all tokenization schemes. We find axes with high overlap between all plural forms (regardless of tokenization scheme) and high discriminability between plural and singular forms, but other axes can still separate different plural tokenization schemes. This work contributes to a growing literature examining the impact of tokenization on the language modeling objective. 
Code and data are available: \url{https://github.com/catherinearnett/spanish-plural-agreement}.

\section{Related Work}

Several studies have investigated morpho-syntactic agreement in BERT-style models across multiple languages \citep[inter alia]{linzen2016assessing, mueller-etal-2020-cross, edmiston2020systematic, perez-mayos-etal-2021-assessing}, finding generally high agreement accuracy. In a subject-verb agreement task, however, BETO incurs a relatively high rate of agreement errors for certain Spanish nouns (despite the ability to extend number agreement to novel words; \citealp{haley2020bert}). It is unclear to what extent degraded performance is attributable to tokenization scheme, but the word ``comanas''---listed as an example of a frequently mis-numbered word---is tokenized non-morphemically into [`coman', `\#\#as'].

Indeed, recent work has demonstrated that morphologically-aware tokenization improves NLP model performance on a variety of downstream benchmarks \citep{park-etal-2020-empirical, hofmann-etal-2021-superbizarre, jabbarmorphpiece, toraman2023impact, uzan2024greed}.
Our work demonstrates how tokenization affects language model predictions involving a specific morphosyntactic rule, providing insight into \textit{how} morphologically-aware tokenization improves NLP model performance.

\section{Model and Data}

All experiments use BETO,  a Spanish pre-trained BERT model \citep{CaneteCFP2020}  with 110M parameters trained on approximately 3B words. 

\subsection{Data}
All plural nouns and their singular form lemmas were extracted from the AnCora Treebanks \citep{alonso2016universal}. Plurals were categorized according to their affix. Nouns ending in vowels use the plural suffix -s, while nouns ending in consonants use the suffix -es. Plurals were also annotated for their grammatical gender by a native Spanish speaker. Irregular nouns, misspellings, and words not listed in the Real Academia Espa\~nola (RAE) online dictionary were excluded.

\subsection{Identifying Tokenization Type}

We created three lists of plurals: one-token ($n$=1247), multi-token morphemic ($n$=508), and multi-token non-morphemic ($n$=627). One-token plurals are stored as single tokens in the tokenizer's vocabulary. We then categorized multi-token plurals as morphemic or non-morphemic. If tokenization followed morpheme boundaries (e.g., \textit{naranjas} as {[}`naranja', `\#\#s'{]}), the noun was categorized as morphemic; if not, it was categorized as non-morphemic (e.g., \textit{neuronas} is tokenized as {[}`neuro', `\#\#nas'{]}).

\subsection{Relationship of Tokenization to Frequency}

Using  oral frequency measures for 2071 target plural wordforms available in a corpus of over 3M spoken words \cite{alonso2011oral}, we examined the relationship between a wordform's frequency and how it was tokenized. A linear model predicting Log Frequency from Tokenization Scheme explained significant variance $[R^2 = 0.33]$. With \textsc{morphemic} level as a reference class (i.e., intercept), the \textsc{non-morphemic} plural nouns were significantly less frequent $[\beta = -0.18, SE = 0.03, p < .001]$, while the \textsc{single-token} plural nouns were significantly more frequent $[\beta = 0.59, SE = 0.03, p < .001]$. As expected, the frequency of a wordform was likely a major factor in how it was tokenized (see also Appendix \ref{subsec:appendix}).

\subsection{Artificial Tokenization Procedure}

To investigate the effect of tokenizing a wordform at the morpheme boundary, we artificially tokenized single-token and multi-token non-morphemic plural nouns by concatenating the token for the appropriate affix (e.g., ``\#\#es'') onto the token(s) for the singular noun (Table \ref{tab:artificial}).

\begin{table}[h!]
\centering
\footnotesize
\begin{tabular}{ccc}
\hline
 \textbf{Morpheme} & \textbf{Original} & \textbf{Artificial} \\ \textbf{Boundary} & \textbf{Tokenization} & \textbf{Tokenization} \\ \hline
 mujer+es                   & {[}`mujeres'{]}                & {[}`mujer', `\#\#es'{]}        \\  
 patrono+s & 
 {[}`patr, `\#\#onos'{]} & 
 {[}`patr, `\#\#ono', `\#\#s'{]} \\ \hline 
\end{tabular}
\normalsize
\caption{Artificial tokenizations for the words \textit{mujeres} `women' (\textit{mujer}), and \textit{patronos} `employers' (\textit{patrono}).}
\label{tab:artificial}
\end{table}

\section{Study: Article-Noun Agreement}
\label{sec:study}

Our primary research question concerned the impact of the original tokenization (\textsc{Tokenization Scheme}) on an article agreement task, similar to that implemented by \citet{linzen2016assessing}.
In Spanish, articles must agree with the \textit{number} of the noun (e.g., \textit{la mujer} vs. \textit{las mujeres}); learned representations for the target noun should thus be conducive to predicting article number. We asked:

\begin{enumerate}
    \item How does the initial tokenization scheme of a plural noun impact the language model's ability to predict the correct article?
    \item Does our \textit{artificial} tokenization scheme provide sufficient information to facilitate successful agreement?
    \item How does the success of our artificial tokenization scheme compare to the original tokenization scheme for those nouns?
\end{enumerate}

\subsection{Method}

Agreement was assessed by taking the logarithm of the relative probability of a plural vs. singular article as predicted by a given noun. For a given wordform (e.g., \textit{mujeres}), a positive log-odds indicated a higher probability was assigned to the plural article, while a negative log-odds indicated a higher probability was assigned to the singular article. A \textit{singular} noun should be associated with a more negative log-odds, while a \textit{plural} noun should be associated with a more positive log-odds. We considered both \textsc{definite} and \textsc{indefinite} articles (\textsc{Article Type}) for each wordform; the log-odds calculation was performed separately for each type.

Accounting for the different presentations of each wordform (i.e., definite vs. indefinite article; original vs. artificial tokenization), our final dataset had 13,276 observations in total, each with an accompanying \textit{log-odds} ratio.
All data and visualizations were analyzed in R; mixed effects models were fit using the \textit{lme4} package \cite{douglas2015fitting}. Maximal random effects structures were fit where possible, and reduced as needed for model convergence.

\subsection{Results}

\subsubsection{Impact of Initial Tokenization}

We first asked whether the original tokenization scheme used for plural nouns affected successful agreement. We fit a mixed model with Log Odds as a dependent variable, fixed effects of Tokenization Scheme and Word Number (and an interaction between the two), fixed effects of Article Type, and random intercepts for each word lemma and sentence. This model explained significantly more variance than a model omitting only the interaction $[\chi^2(2) = 6.54, p = .04]$, suggesting that different tokenization schemes were differentially successful in predicting the appropriate article. Note that this interaction was independent from the effect of wordform frequency (see Appendix \ref{subsec:appendix}).

\begin{figure}[h!]
    \centering
    \includegraphics[width=\linewidth]{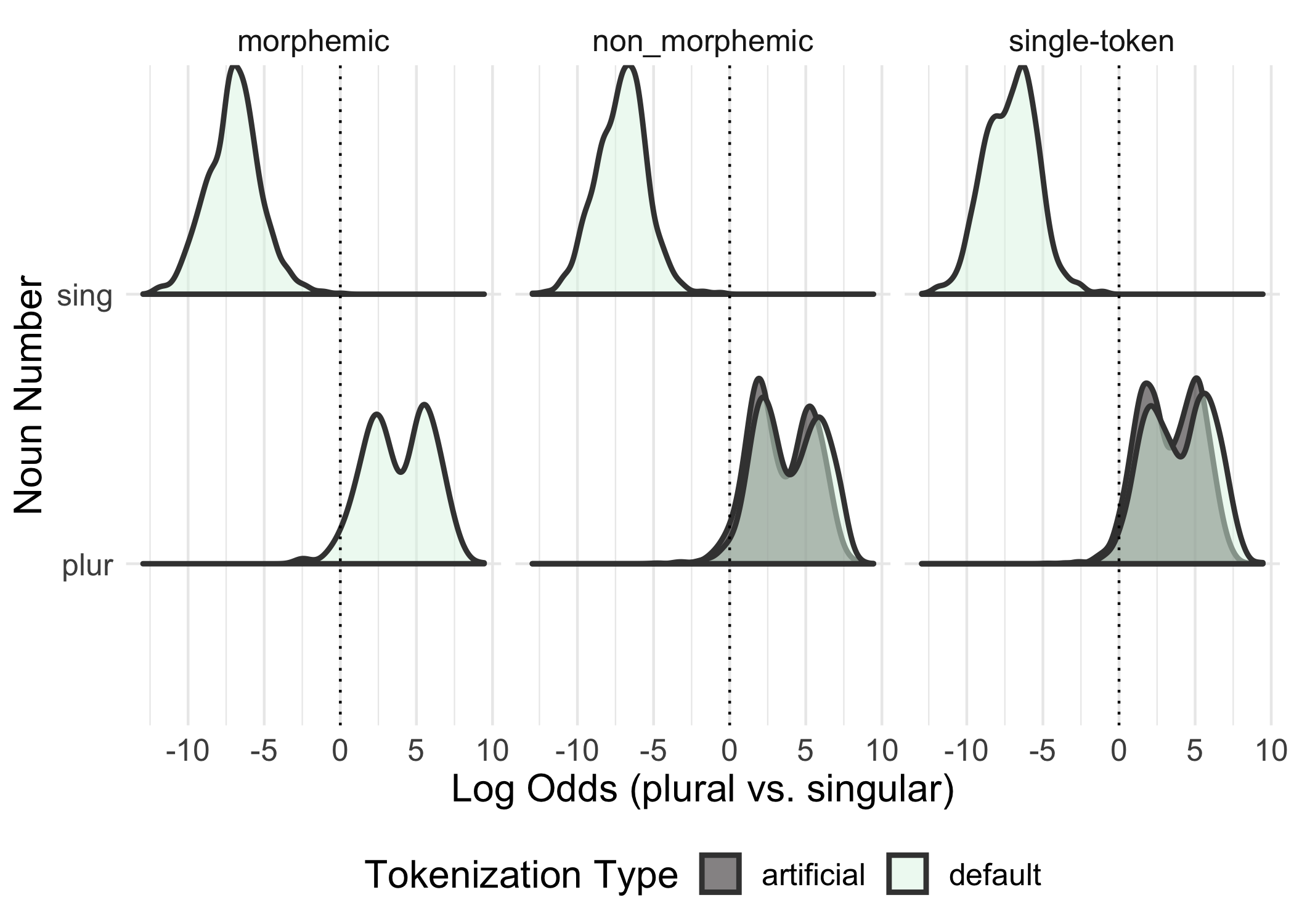}
    \caption{Log-odds varied significantly as a function of noun number (\textit{singular} vs. \textit{plural}). The extent of this variance interacted (weakly) with initial tokenization (\textit{morphemic} vs. \textit{non-morphemic} vs. \textit{single-token}) and with whether the \textit{original} or \textit{artificial} tokenization procedure was used. Larger log-odds indicate higher probabilities of the plural article.}
    \label{fig:density_figure}
\end{figure}

However, as depicted in Figure \ref{fig:density_figure}, this effect was quite small. Accuracy was near ceiling for all tokenization types, i.e., the Log Odds was larger than 0 for plural nouns and smaller than 0 for singular nouns (see also Table \ref{tab:accuracy_original}). Thus, our results do not suggest that morphologically-aligned tokenization is required for good agreement performance.

\begin{table}[h!]
\centering
\begin{tabular}{ccc}
\hline
 \textbf{Original Tokenization} & \textbf{Original} &
 \textbf{Artificial}\\ \hline
 Morphemic                  & 
 0.97  &
 ---\\  
 Non-morphemic & 
 0.98 &
 0.96\\ 
 Single-Token                  & 
 0.98    &
 0.97 \\  \hline 
\end{tabular}
\caption{Accuracy scores for \textit{plural nouns} only, using either the original tokenization scheme for that class of nouns or the artificially-induced morphemic scheme.}
\label{tab:accuracy_original}
\end{table}

\subsubsection{Success of Artificial Tokenization}

Next, we artificially tokenized plural nouns that would otherwise be tokenized non-morphemically or as a single-token. To quantify the success of this procedure, we fitted a linear mixed-effects model predicting Log Odds with fixed effects of Article Type, Word Number, Tokenization Scheme, and Affix (`\#\#s' or ``\#\#es''), as well as random intercepts for word lemma and sentence. 

This model explained significantly more variance than a model omitting only Word Number $[\chi^2(1) = 11988, p < .001]$, indicating that the artificial tokenization procedure still led to good article number agreement performance: Log Odds were significantly different for singular nouns and artificially-tokenized plural nouns (see also Figure \ref{fig:density_figure} and Table \ref{tab:accuracy_original}).

\subsubsection{Comparing Default vs. Artificial Tokenization Schemes}

Finally, restricting our analysis to plural forms, we asked whether a higher Log Odds was assigned to \textit{artificially tokenized} plural nouns than ones using the default scheme. We fitted a linear mixed-effects model with fixed effects of Tokenization Scheme (artificial or original), Affix, and Original Tokenization Scheme (as well as random intercepts for word lemma, sentence, and wordform, and by-lemma random slopes for Tokenization Scheme). This model did explain more variance than a model omitting only Tokenization Scheme $[\chi^2(1) = 141.81, p < .001]$. Critically, however, the Log Odds for the artificially tokenized plural nouns was \textit{lower} ($M = 3.38, SD = 2$) than when using the default tokenization ($M = 3.95, SD = 2.15$). In other words, the artificially-induced morphemic tokenization was successful, but less so than relying on the original scheme for those nouns.

\section{Linear Discriminant Analysis (LDA)}
\label{sec:lda}
\begin{figure}[h!]
    \centering
    \includegraphics[width=0.9\linewidth]{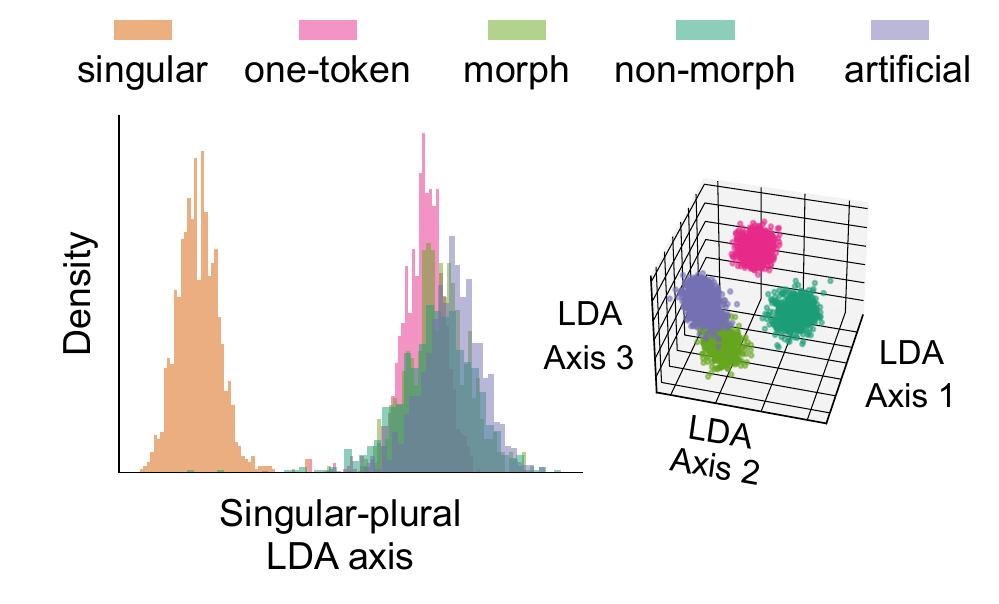}
    \caption{LDA for singular and plural embeddings reveals axes of overlap (\textit{left}) and discriminability (\textit{right}) for differentially tokenized plural forms.}
    \label{fig:lda}
\end{figure}
To identify potential causes for the observed agreement patterns across noun types (singular vs. different plural tokenizations), we considered the embeddings of those nouns in the language model representation space.
We took each noun's mean embedding across the last four (out of twelve) BETO Transformer layers, averaging over all tokens in the noun.
To minimize confounds from averaging embeddings over different numbers of tokens, we considered only two-token plurals in all multi-token scenarios for embedding analyses.

We first identified the linear axis that maximally separated single-token singular from plural nouns.
To do this, we ran linear discriminant analysis (LDA) with two classes of embeddings: singular nouns (all single-token) and single-token plural nouns.\footnote{Given $n$ sets of representations, LDA computes $n-1$ directions in the language model representation space that maximize separation between the sets.}
We then projected all noun representations linearly onto this axis, essentially projecting each embedding into a single value.
As expected, we found that singular nouns clustered separately from plural nouns (Figure \ref{fig:lda}, \textit{left}).
Notably, all types of plurals (single-token, artificially tokenized, two-token morphemic, and two-token non-morphemic) patterned together and were not linearly discriminable along this axis.
This suggests that the model could rely on similar number agreement mechanisms for different types of plurals, but future work would need to demonstrate causal impacts of this singular-plural axis on number agreement predictions (e.g. as in \citealp{mueller-etal-2022-causal}).

While the singular-plural LDA axis mapped different plural types to similar values, other axes could separate embeddings for the different plural types.
We used LDA to identify the three linear axes that maximally separated the four types of plurals. As shown in Figure \ref{fig:lda} (\textit{right}), single-token plurals and two-token non-morphemic plurals were separable from one another and from all other plural types.
The artificial and default morphemic plurals had distinct clusters, but they were not entirely separable from one another.
This indicates that even though the artificial tokenization was never seen by the model during training, the representations were still quite similar (e.g. due to the presence of the `\#\#s' or `\#\#es' token).
The slight separation between these clusters may be driven either by frequency effects or by veridical differences in how the models represent number in the two plural types.

\section{Discussion and Conclusion}
\label{sec:discussion}

We assessed whether distinct tokenization schemes impacted the ability of BETO (a Spanish language model) to predict appropriate \textit{articles} for Spanish plural nouns. Single-token representations facilitated slightly better predictions overall. However, the model did show evidence of \textit{generalization} consistent with having learned morpheme-like ``rules'': artificially re-tokenizing plural nouns along morpheme boundaries produced representations amenable to article prediction---despite the language model never having previously observed that sequence of tokens (see Figure \ref{fig:density_figure})---though this approach was slightly less accurate than relying on the original tokenization scheme. This provides further insight into work on language models generalizing morphological patterns \citep{haley2020bert}; however, this does not work equally well for all languages or models \citep{weissweiler2023counting}.

Notably, the similar agreement performance across single-token, morphological, non-morphological, and artificially-tokenized plurals could indicate multiple different agreement mechanisms in the model.
Future work might apply causal interventions on different embedding axes (as found in \S\ref{sec:lda}), to determine the extent to which the same model subnetworks are involved in number agreement for different types of plural tokenizations, shedding light on the impacts of tokenization on language model processing.

\section{Limitations}

A key limitation of the current work is scope. Future work could consider additional morphological phenomena, additional languages, and a larger range of language models or tokenization schemes.
A second limitation is that the language model's performance was near-ceiling for each category considered. Future work could work to develop more challenging tasks for which the model is not at ceiling (as in \citealp{linzen2016assessing}).
Finally, our work does not demonstrate the extent to which different tokenizations rely on the same internal mechanisms for agreement in the model (\S\ref{sec:discussion}), which is a valuable direction for future work.

\section{Acknowledgements}
Tyler Chang is partially supported by the UCSD Halıcıoğlu Data Science Institute graduate fellowship, and Pamela D. Rivi\`ere is supported by UCSD's Chancellor's Postdoctoral Fellowship Program.

\bibliography{anthology,custom}
\bibliographystyle{acl_natbib}

\appendix
\section{Appendix}
\label{sec:appendix}

\subsection{Article Agreement Task: Model Inputs}

\begin{itemize}
\item Example model inputs for original single-tokenizations: ``[CLS] [MASK] mujeres [SEP]''

\item Example model inputs for artificial (morphemic) tokenizations: ``[CLS] [MASK] mujer \#\#es [SEP]''

\item Example model inputs for original non-morphemic multi-tokenizations: ``[CLS] [MASK] patr \#\#onos [SEP]''

\item Example model inputs for artificial (morphemic) tokenizations: ``[CLS] [MASK] patr \#\#ono \#\#s [SEP]''
\end{itemize}

\subsection{Supplementary Analysis with Log Frequency}\label{subsec:appendix}

\begin{figure}[h!]
    \centering
    \includegraphics[width=\linewidth]{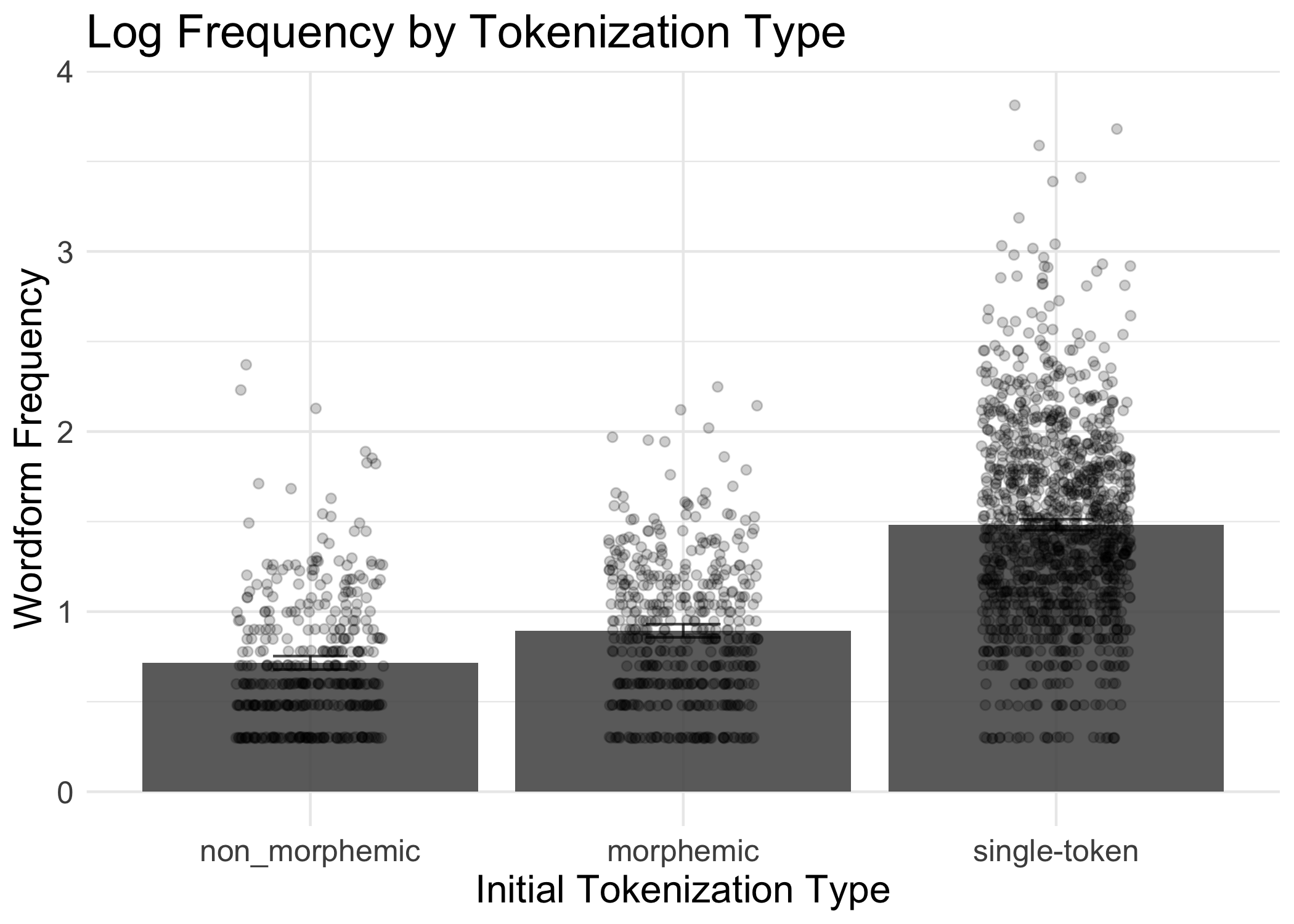}
    \caption{Single-token plurals were significantly more frequent than those tokenized according to morphemic boundaries, which were more frequent than those tokenized according to non-morphemic substrings.}
    \label{fig:freq_tokenization}
\end{figure}

We ran a follow-up analysis asking whether the Log Frequency of a wordform was predictive of agreement success. This analysis had two key goals. First, because Log Frequency was correlated with Tokenization Scheme, we aimed to determine whether the effect of Tokenization Scheme on agreement success was in fact due to effects of token frequency. Second, we were independently interested in whether the language model made better predictions for more frequent wordforms.

We fitted a linear mixed-effects model including fixed effects of Tokenization Scheme, Word Number, and Log Frequency, as well as interactions between Word Number and Tokenization Scheme and between Word Number and Log Frequency. We also included random intercepts for word lemma and sentence. This model explained significantly more variance than a model omitting only the interaction between Log Frequency and Word Number $[\chi^2(1) = 17.89, p < .001]$. The interaction was negative $[\beta = -0.35, SE = 0.08, p < .001]$, i.e., the plural article log-odds were \textit{more} negative for more frequent singular nouns. In other words, the language model made better predictions for more frequent nouns than less frequent nouns.

The full model also explained more variance than a model omitting the interaction between Word Number and Tokenization Scheme $[\chi^2(2) = 11.24, p = .004]$. This indicates that even controlling for wordform frequency, there was an independent effect of how the wordform was initially tokenized on the success of the language model's article predictions.

\end{document}